\documentclass[english,a4paper,11pt]{article}

\usepackage{amsmath, amsxtra, amsfonts, amssymb, amstext, amsthm}
\usepackage{booktabs}
\usepackage{graphics}
\usepackage{babel}
\usepackage[noadjust]{cite}
\usepackage{url}
\usepackage[algo2e,ruled,vlined,linesnumbered]{algorithm2e}
\usepackage{todonotes}
\usepackage{fullpage}

\usepackage{nicefrac}
\usepackage[colorlinks]{hyperref}
\definecolor{linkblue}{rgb}{0.1,0.1,0.8}
\hypersetup{colorlinks=true,linkcolor=linkblue,filecolor=linkblue,urlcolor=linkblue,citecolor=linkblue}
\usepackage{wrapfig}

\newcommand{\N}{\mathbb{N}}
\newcommand{\R}{\mathbb{R}}

\renewcommand{\epsilon}{\varepsilon}

\newcommand{\disc}{d_{\infty}^*}

\begin{document}

\title{Constructing Low Star Discrepancy Point Sets with Genetic Algorithms}

\author{Carola Doerr$^1$, Fran\c cois-Michel De Rainville$^{2}$}
\date{$^1$LIAFA, Universit\'e Paris Diderot - Paris 7, 75205 Paris Cedex 13, France and\\
Max Planck Institute for Informatics, 66123 Saarbr\"ucken, Germany\\[1ex]
$^2$Laboratoire de vision et syst\`emes num\'eriques\\ 
D\'epartement de g\'enie \'electrique et de g\'enie informatique, Universit\'e Laval\\
Qu\'ebec, Qu\'ebec, Canada~~G1V 0A6
}

\maketitle

\maketitle
\begin{abstract}
Geometric discrepancies are standard measures to quantify the irregularity of distributions. They are an important notion in numerical integration. 
One of the most important discrepancy notions is the so-called \emph{star discrepancy}. Roughly speaking, a point set of low star discrepancy value allows for a small approximation error in quasi-Monte Carlo integration. It is thus the most studied discrepancy notion. 

In this work we present a new algorithm to compute point sets of low star discrepancy. The two components of the algorithm (for the optimization and the evaluation, respectively) are based on evolutionary principles. 
Our algorithm clearly outperforms existing approaches. 
To the best of our knowledge, it is also the first algorithm which can be adapted easily to optimize inverse star discrepancies.
\end{abstract}

\textbf{Categories:}
{F.2.1} {Theory of Computation}{Analysis of Algorithms and Problem Complexity} [Numerical Algorithms and Problems]\\
{I.2.8} {Artificial Intelligence}{Problem Solving, Control Methods, and Search} [Heuristic Methods]\\


\section{Introduction}

\begin{figure}[t]
\begin{center}
\includegraphics[scale=.9]{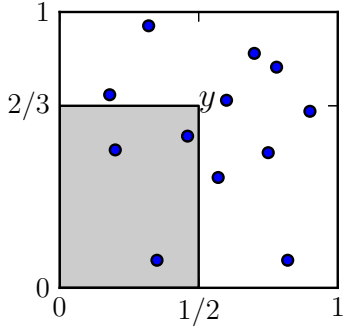}
\end{center}
\caption{The local discrepancy of $X$ in $y$ is $2/3 \cdot 1/2-3/12=1/12$.}
\label{fig:localstardisc}
\end{figure}

For a point set $X=\{x^{(1)}, \ldots, x^{(n)}\}\subset [0,1)^d$ and a point $y$ in $[0,1]^d$, the local star discrepancy $\disc(X,y)$ of $X$ in $y$ measures the absolute difference between the volume of the box $[0,y)$ and the fraction of points of $X$ that are inside that box, cf.~Figure~\ref{fig:localstardisc}. 
The star discrepancy of $X$ is the largest such value; i.e., 
$\disc(X):=\sup_{y \in [0,1]^d}{\disc(X,y)}$.

Star discrepancies are deeply related to the ubiquitous task of numerical integration but have found several applications also in computer graphics, pseudorandom number generators, experimental design, and option pricing, to name but a few domains.
The approximation error of Monte Carlo and quasi-Monte Carlo methods can be expressed in terms of the star discrepancy. To be more precise, let $f:[0,1]^d \rightarrow \R$ be a function for which we want to compute the integral $\int_{[0,1]^d}{f(s){\rm d}s}$. The Koksma-Hlawka inequality~\cite{Hla61, Nie92} states that, for suitable $f$, the difference between this integral and the average function value of $f$ on a point set $X$ is bounded by 
\begin{equation}
\left| \int_{[0,1]^d}{f(s) {\rm d}s} - \frac{1}{n}\sum_{i=1}^n{f(x^{(i)})} \right|
\leq
V(f)\cdot \disc(X)\,,
\label{eq:koksma}
\end{equation}
where $V(f)$ denotes the so-called variation in the sense of Hardy and Krause. Since $V(f)$ depends only on $f$ and not on $X$, the smaller the discrepancy $\disc(X)$, the better approximation we can expect.

While the points $x^{(1)}, \ldots, x^{(n)}$ in Eq.~\ref{eq:koksma} are chosen randomly in Monte Carlo integration, it has been known for a long time that we can achieve better results by evaluating $f$ in \emph{low discrepancy point sets}. 
Thus, in quasi-Monte Carlo numerical integration, the set $X=\{x^{(1)}, \ldots, x^{(n)}\}$ is chosen deterministically, so that the factor $\disc(X)$ in Eq.~\ref{eq:koksma} is as small as possible. 
It is thus one of the main challenges in numerical analysis to compute explicit point sets $X$ of smallest possible star discrepancy value. 

\subsection{Scope and Previous Related Work}
In this work, we present a new algorithm for constructing low discrepancy point sets. 
In particular, we provide improved upper bounds for the following two questions. 
\begin{enumerate}
	\item[1] \emph{Star discrepancy:} For given $n$ and $d$, what is the smallest star discrepancy that can be achieved by an $n$-point configuration in $[0,1)^d$? 
	\item[2] \emph{Inverse star discrepancy:} For given $d$ and $\varepsilon$, what is the smallest possible value of $n$ such that there exists a set $X \subset [0,1)^d$ of size $|X|=n$ such that $\disc(X) \leq \varepsilon$?
\end{enumerate}

Note that the first question asks where to place the $n$ points such that the irregularity measured by the star discrepancy is as small as possible, while the second question is even more involved. It asks us to decide how many points we need and (indirectly) where to place them, such that the resulting irregularity is at most $\varepsilon$. 

Our algorithm has originally been developed to address the first question. However, it turns out that, using a bisection approach and multi-objective selection (see Section~\ref{sec:algorithm}), we can easily adapt it to answer also the second question. 

Our algorithm builds on previous work presented in~\cite{DeRainville2012} on the construction of low $L_2$-discrepancy sequences (see~\cite{DeRainville2009} for an earlier GECCO version) and in~\cite{GWW12} on the evaluation of the star discrepancy of a given point set.
It has two components, an \emph{optimization component}, which computes candidate point sets, and an \emph{evaluation component} for assessing the quality of the proposed solutions. During the optimization process, the evaluation component is called several times. 
The two components of the algorithm will be described in Section~\ref{sec:algorithm}. On the structure of the algorithms we note here only that the optimization part is based on a genetic algorithm, while the evaluation component is based on threshold accepting (TA)---a variant of simulated annealing with derandomized selection rules.

Evaluating the star discrepancy of a given point set $X$ is known to be NP-hard~\cite{GSW09}. In fact, it is even W[1]-hard in $d$~\cite{GKWW12}, implying that, under standard complexity assumptions, there is no algorithm to evaluate the star discrepancy of $n$ points in $d$ dimension in a running time $n^{o(d)}$. The best known exact algorithm for evaluating discrepancies, the DEM-algorithm, has a running time of $n^{1+d/2}$~\cite{DEM96}. For most relevant settings this is too slow to be applicable. This is true in particular for our setting, where  many candidate point sets need to be evaluated. In fact, the complexity of star discrepancy evaluation is the main reason why only few algorithmic approaches are known for the explicit construction of low star discrepancy point sets, cf.~also the comment in~\cite[page 3]{DeRainville2012}.

A new robust algorithm to estimate star discrepancy values has been proposed in~\cite{GWW12}. This algorithm has been reported to give very accurate discrepancy estimates, and our experiments confirm these statements. We thus use this algorithm for the intermediate discrepancy evaluations; i.e., for the optimization process of creating good candidate point configurations. Where feasible, we do a final evaluation of the candidate sets using the exact DEM-algorithm described above. 

\subsection{Structure of the Paper}
An introduction to the star discrepancy problem is given in Section~\ref{sec:stardisc}. We discuss there also the issue of computing star discrepancy values, and we introduce the generalized Halton sequence, for which our algorithm finds good generating vectors. The algorithm itself is presented in Section~\ref{sec:algorithm}. 

Section~\ref{sec:results} surveys our empirical results. For the star discrepancy problem we compare our results with those presented in~\cite{DGW10, Thi01JoC, Thi01b}. The point sets computed by our algorithm clearly outperform those reported there. We also show that our algorithm produces surprisingly good bounds for the inverse star discrepancy problem, thus easily answering one of the subproblems in Open Question 42 from~\cite{NW10}. Our bounds for all three subproblems are better by a factor~(!) of 5 to 7 compared to what was asked for in~\cite{NW10}, but for two subproblems our values are computed only by the algorithms from~\cite{GWW12} and need to be verified by computations with the DEM-algorithm.

\section{Star Discrepancies}
\label{sec:stardisc}

Throughout this work, we denote by $n$ the size of the set $X:=\{x^{(1)}, \ldots, x^{(n)}\} \subset [0,1)^d$ under consideration and by $d$ the dimension of the problem instance. 

For a point $y \in [0,1]^d$ the local discrepancy of $y$ with respect to $X$ is defined by
\begin{align*}
	\disc(y,X):=\left| V_y - \frac{A(y,X)}{n}\right|\,,
\end{align*}
where $V_y:=\prod_{j=1}^d{y_j}$ denotes the Lebesgue volume of the box $[0,y)$ and \begin{align*}
A(y,X):=|\{ i \in \{1,\ldots,n\} \mid \forall j \in \{1,\ldots,d\}: x^{(i)}_j < y_j \}|
\end{align*}
is the number of points of $X$ that fall into this box. 
The discrepancy of $X$ is 
\begin{align*}
	\disc(X):=\sup_{y \in [0,1]^d} {\disc(y,X)}\,,
\end{align*}
the largest local discrepancy value. 

\subsection{Computation of Star Discrepancies}
\label{subsec:computation}

\begin{figure}[t]
\begin{center}
\includegraphics[scale=.9]{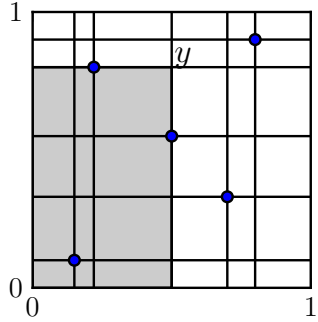}
\end{center}
\caption{The discrepancy of $X$ is obtained in one of the grid points $y \in \Gamma(X)$.}
\label{fig:grid}
\end{figure}

It has been observed in~\cite{Nie72a} that the computation of $\disc(X)$ can be discretized. In fact, if we let 
$\Gamma_j(X):=\{ x^{(i)}_j \mid 1 \leq i \leq n \}$, the set of $X$-values in the $j$th coordinate ($1 \leq j \leq d$), 
and $\Gamma(X):=\prod_{1 \leq j \leq d}{\left(\Gamma_j(X) \cup \{1\}\right)}$ be the \emph{grid spanned by} $X$, 
then 
\begin{align*}
	\disc(X):=\max_{y \in \Gamma(X)} \left\{ V_y - \frac{A(y,X)}{n}, \frac{\bar{A}(y,X)}{n}-V_y \right\}\,,
\end{align*}
where $\bar{A}(y,X):=|\{ i \in \{1,\ldots,n\} \mid \forall j \in \{1,\ldots,d\}: x^{(i)}_j \le y_j \}|$ is simply the number of points of $X$ that fall into the \emph{closed} box $[0,y]$. 
That is, instead of evaluating the continuous space $[0,1]^d$, the search space can be reduced to a discrete one of size $(n+1)^d$, cf.~Figure~\ref{fig:grid}. 

However, for most practical purposes, $2(n+1)^d$ function evaluations are way too many, and one has to resort to different methods for evaluating star discrepancy values. As mentioned in the introduction, computing the star discrepancy of a given point set is known to be a hard problem, and the best known exact algorithm, the DEM-algorithm from~\cite{DEM96}, has a running time of $n^{1+d/2}$. A new promising algorithm for \emph{approximate} star discrepancy evaluation has been presented in~\cite{GWW12}. We discuss this algorithm in Sec.~\ref{sec:fitnessevaluation}.

Given the importance of low star discrepancy point sequences in numerical integration and its various other applications in computer science, it is thus mainly due to the complexity of evaluating star discrepancies that not much work has been done in the search heuristics community to construct low star discrepancy point sequences. On the other hand, there has been some effort to construct low discrepancy point sets for other measures of irregularity, see, for example, the work in~\cite{DeRainville2012}, which constructs point sets that are optimized for \emph{Hickernell's modified $L_2$-discrepancy}---a measure that can be computed efficiently in $O(dn^2)$ arithmetic operations. 
See~\cite{DGW13} for a recent survey on the computation of geometric discrepancies.

\subsection{Low Star Discrepancy Point Sets}

The ultimate goal of star discrepancy theory is to find, for given parameters $n$ and $d$, a set $X \subset [0,1)^d$ of size $|X|=n$ such that $\disc(X)$ is as small as possible. 
A closely related question of similar practical and theoretical relevance is the \emph{inverse star discrepancy} problem: for given $d$ and $\varepsilon>0$, what is the smallest integer $n$ such that there exists a point set $X$ of size $|X|=n$ satisfying $\disc(X) \leq \varepsilon$.

Since the early twentieth century, a lot of work has been done to address these two questions, see~\cite{DP10,Hinrichs13,NW10,Gne12b} for recent surveys. Many different low discrepancy sequences have been defined (e.g., Sobol, Faure, and Halton sequences), and theoretical bounds on their performance have been proven. 
Furthermore, general lower bounds for the star discrepancy of any $n$-point set in $d$ dimensions exist. 
However, the problem with all these bounds (in the interest of space, we cannot give a detailed survey here but we refer to~\cite{Gne12b} for a concise summary of the known bounds) is that they are either asymptotic statements (and thus of limited relevance in the regime of practical interest) or the precision of the results is not accurate enough to allow for a comparison between the different constructions. 
We note also that there is a gap between all known lower bounds and the best known upper bounds for the star discrepancy problem. 
It thus remains a challenging open question to construct such point sets, and it is a well suitable problem for demonstrating the strength of bio-inspired approaches for classical optimization challenges.

In~\cite{DeRainville2012}, the candidates for the $L_2$-optimized point sets are so-called \emph{generalized Halton sequences}. Since these point sets exhibit not only low $L_2$ discrepancies, but also low star discrepancies, we use here the same construction for finding our candidate sets. Given the lack of space, we give here only the required definitions of the point sets, and we refer the interested reader to~\cite{DeRainville2012} and the books~\cite{Mat99,Tez95} or the recent survey~\cite{DGW13} for further information on generalized Halton sequences. 

\textbf{The Generalized Halton Sequence.}
The Halton sequence~\cite{Halton1960} is a generalization of the one-dimensional \emph{van der Corput sequence}~\cite{vanderCorput1935}. 
For any prime number $p \in \N$ the $i$th number of this latter sequence in base $p$, is given by
\begin{align}
	\varphi_p(i) := \sum_{\ell=1}^k d_{\ell} p^{-\ell}\,,
	\label{eqn:VDC}
\end{align}
where $i=d_k d_{k-1} \ldots d_1$ is the digital expansion of $i$ in base $p$; i.e., $i = \sum_{\ell=1}^k d_{\ell} p^{\ell-1}$.

The multidimensional Halton sequence is simply obtained by grouping together van der Corput sequences of different bases. 
More precisely, let $p_1, \ldots, p_d$ be the first $d$ prime numbers. 
The $i$th element of the $d$-dimensional Halton sequence is given by
\begin{align*}
	x^{(i)} := (\varphi_{p_1}(i), \ldots, \varphi_{p_d}(i))\,.
\end{align*}

The Halton sequence is a low-discrepancy sequence; i.e., its first $n$ points
$X=(x^{(i)})^n_{i=1}$ in dimension $d$ satisfy the star discrepancy bound
\begin{align}
\label{halton_bound}
 d^*_\infty(X) = O \big( n^{-1}\,\ln(n)^d \big).
\end{align}
In fact, the Halton sequence was the first construction for which Eq.~\ref{halton_bound} 
was verified for any dimension $d$~\cite{Halton1960}, and up to now there is no sequence
known that exhibits a better asymptotic behavior.

Nevertheless, the Halton sequence suffers from strong correlation between the sequences in high dimensions, thus motivating Braaten and Weller~\cite{Braaten1979} to suggest a \emph{generalized} (also referred to as \emph{scrambled}) \emph{Halton sequence} .
In this sequence, Eq.~\ref{eqn:VDC} is replaced by
\begin{equation*}
	\varphi^{\pi_p}_p(i) :=\sum_{\ell=1}^k{\pi_p(d_{\ell})\,p^{-\ell}}\,,
	\label{eq:ScrambledVDC}
\end{equation*}
where $\pi_p$ is a permutation of $\{0,1,\ldots,p-1\}$ with fixpoint $\pi_p(0)=0$ (so that $\varphi^{\pi_p}_p(i) \neq 0$).
The $i$th element of the $d$-dimensional generalized Halton sequence is then defined by
\begin{align}
\label{eqn:ghalton}
x^{(i)}(\boldsymbol{\Pi}):=(\varphi^{\pi_{p_1}}_{p_1}(i), \ldots, \varphi^{\pi_{p_d}}_{p_d}(i))\,,
\end{align}
where we abbreviate $\boldsymbol{\Pi}:=(\pi_{p_1}, \ldots, \pi_{p_d})$.

To answer the discrepancy question, we thus need to find a vector of permutations $\boldsymbol{\Pi}$ such that the star discrepancy of the point set $X(\boldsymbol{\Pi})=\{x^{(1)}(\boldsymbol{\Pi}), \ldots, x^{(n)}(\boldsymbol{\Pi})\}$ is as small as possible.
Since the point set is completely determined by the permutations $\boldsymbol{\Pi}$, we call $\boldsymbol{\Pi}$ the \emph{generating vector of} $X(\boldsymbol{\Pi})$.

\section{Algorithm}
\label{sec:algorithm}

The optimization of the generating vectors for the generalized Halton sequence is made with a very simple genetic algorithm. As presented in Algorithm \ref{algo:opt}, a $(\mu+\lambda)$ scheme is used. In each generation an offspring population $\mathfrak{P}_\text{o}^{(g)}$ of $\lambda$ individuals is generated from the parental population $\mathfrak{P}_\text{p}^{(g)}$. Each individual is produced using either mutation or crossover according to some probabilities. The new individuals are evaluated ($f(X(\boldsymbol{\Gamma}_i))$ in line~\ref{line:evaluation}) and the parents reevaluated ($f^\prime(X(\boldsymbol{\Pi}_i))$ in line~\ref{line:reevaluation}) with respect to the discrepancy of the point set that they generate. The selection of the next generation parental population is made based on the fitness of the individuals of both populations $\mathfrak{P}_\text{p}^{(g)}$ and $\mathfrak{P}_\text{o}^{(g)}$. We use as stopping criterion a fixed number of generations. This section describes in more detail each component of the genetic algorithm.

\begin{algorithm2e}[t]
	initialize $\mathfrak{P}_\text{p}^{(1)} \leftarrow \left\lbrace(\boldsymbol{\Pi}_i, f(X(\boldsymbol{\Pi}_i))) \mid i = 1, \ldots, \mu\right\rbrace$\;
	\While{$\neg$stop}{ \label{line:while}
		\For{$i= 1, \ldots, \lambda$}{
			\uIf{crossover}{
				$\boldsymbol{\Pi}_1, \boldsymbol{\Pi}_2 \leftarrow \operatorname{select\_random}\left(\mathfrak{P}_\text{p}^{(g)}, 2\right)$\;
				$\boldsymbol{\Gamma}_i \leftarrow \operatorname{mate}\left(\boldsymbol{\Pi}_1, \boldsymbol{\Pi}_2\right)$\;
			}
			\ElseIf{mutation}{
				$\boldsymbol{\Pi} \leftarrow \operatorname{select\_random}\left(\mathfrak{P}_\text{p}^{(g)}, 1\right)$\;
				$\boldsymbol{\Gamma}_i \leftarrow \operatorname{mutate}\left(\boldsymbol{\Pi}\right)$\;
			}
		}
		$\mathfrak{P}_\text{o}^{(g)} \leftarrow \left\lbrace\left(\boldsymbol{\Gamma}_i, f(X(\boldsymbol{\Gamma}_i))\right) \mid i = 1, \ldots, \lambda\right\rbrace$\label{line:evaluation}\;
		$\mathfrak{P}_\text{p}^{(g)} \leftarrow \left\lbrace\left(\boldsymbol{\Pi}_i, \max\{f(X(\boldsymbol{\Pi}_i)), f^\prime(X(\boldsymbol{\Pi}_i))\}\right) \mid i = 1, \ldots, \mu\right\rbrace$\label{line:reevaluation}\;
		$\mathfrak{P}_\text{p}^{(g+1)} \leftarrow \operatorname{select}\left(\mathfrak{P}_\text{p}^{(g)} \cup \mathfrak{P}_\text{o}^{(g)}, \mu\right)$\;
	}
	\caption{Genetic Optimization}
	\label{algo:opt}
\end{algorithm2e}

\subsection{Representation}
Following a similar procedure as in \cite{DeRainville2012}, we optimize over the generating vector $\boldsymbol{\Pi}$ of the generalized Halton sequence. The difference of our work compared to \cite{DeRainville2012} is the objective function---\cite{DeRainville2012} considers $L_2$-discrepancies while we optimize for star discrepancy---and the fact that here in our work, we are interested in generating low star discrepancy \emph{point sets}, not \emph{sequences}. This allows us to do the optimization for all the permutations at the same time (since we do not need to find a configuration that would be good also for all smaller number of points).
Therefore, we use an adjusted implementation of the algorithm from~\cite{DeRainville2012}. 

\begin{figure}
	\centering
	\includegraphics[scale=.9]{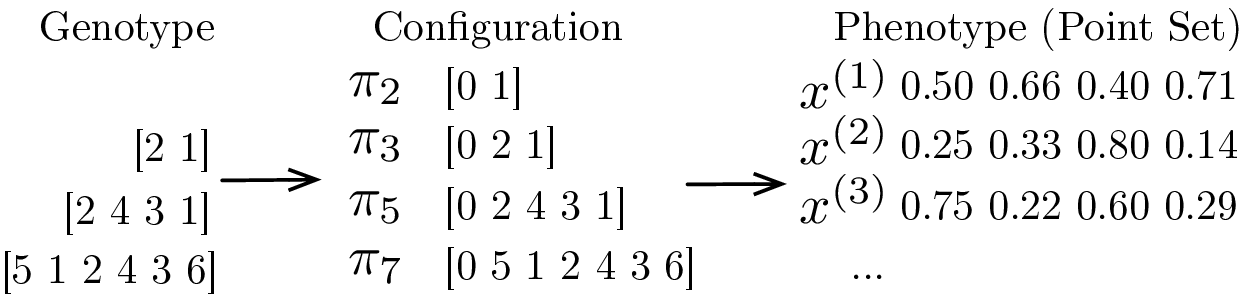}
	\caption{Representation of the individual's genotype and its conversion to the phenotype.}
	\label{fig:representation}
\end{figure}

For the optimization of a $d$-dimensional point set, the genotype of the individuals contains $d-1$ independent permutations.
By definition, the first permutation of the generating vector is always $[0~1]$ since the first prime number $p_1$ is 2 and we require that $\pi_i(0) = 0$ always. For the same reasons, the second permutation $\pi_3$ is either $[0~1~2]$ or $[0~2~1]$ and the third one, $\pi_5$ is chosen from all $4!$ possible permutations of $\{0,1,\ldots, 4\}$ with $\pi_5(0)=0$.
For technical reasons, the $0$ is removed from the permutations in the genotype, and we call the permutation itself (with the prepending 0) the \emph{configuration}.
Finally, the point set $X$ is generated by Eq.~\ref{eqn:ghalton} with the generating vector $\boldsymbol{\Pi}$ set to the configuration constructed from the genotype.
An example showing the translation from a genotype into a phenotype for a 4-dimensional point set is given in Figure~\ref{fig:representation}.

\subsection{Variations}
As seen in Algorithm~\ref{algo:opt}, crossover and mutation are used to produce the offspring. These operators, when applied on an individual, affect all underlying permutation vectors of its genotype. The crossover chosen is the partially matched crossover \cite{Goldberg1985}; it chooses two crossover points and exchanges the alleles between these two points. The mutation is a uniform partial reordering of the alleles as presented in \cite{DeRainville2012}; the reordered alleles are chosen by a uniform matching probability. Both operators preserve the validity of a permutation representation; i.e., the resulting offspring are again permutations of the required length.

\subsection{Fitness Evaluation}\label{sec:fitnessevaluation}

Naturally, the fitness of the individuals is the star discrepancy of the point set that they generate; i.e., the star discrepancy of their phenotype. We aim at minimizing this value.
To assess the discrepancy values we use two different algorithms, the DEM-algorithm proposed in~\cite{DEM96} and the TA-algorithm from~\cite{GWW12}.
 
The DEM-algorithm is an exact algorithm for computing the star discrepancy of a given point set. It is based on dynamic programming and has a running time of $n^{1+d/2}$. 
The algorithm has been implemented by M. Wahlstr\"om and is available from~\cite{Wahl12}. We use his implementation to evaluate point sets for up to 9 dimensions. 
For these point sets, the reevaluation step in line~\ref{line:reevaluation} of Algorithm~\ref{algo:opt} can be skipped. 

Where the exact DEM-algorithm has an infeasible running time, we resort to the TA-heuristic proposed in~\cite{GWW12}. This algorithm gives a lower bound for the discrepancy of a point set and we use this lower bound to guide our optimization procedure. 

In the interest of space, we cannot give a full description of this TA-approach but provide only a high level overview. The goal of the TA-algorithm is to find a good (i.e., large) lower bound for the star discrepancy of the given point set $X$. To this end, it performs a guided random search on the grid spanned by $X$, cf.~Section~\ref{subsec:computation}. 
The algorithm uses a $(1+1)$ scheme; i.e., it keeps one individual at a time.
In each iteration, an offspring is created by sampling according to some probability distribution from the neighborhood of the parent individuum. The local star discrepancy is computed, and the offspring replaces the parent if its local discrepancy value is larger than or at least not much smaller than the parent's one. The ``not much smaller part'' is quantified by a threshold value $T<0$; i.e., we replace the parent grid point $p$ by the offspring point $o$ if and only if $\disc(o,X)-\disc(p,X) \ge T$. The threshold value changes over time, and converges to $0$ so that the algorithm finally outputs a local maximum of the star discrepancy value of $X$.

In order to gain precision on our estimation, the TA-algorithm is reapplied on the parental population in each generation (line~\ref{line:reevaluation}) and the individual's fitness is the maximum value of the current estimate and the newly calculated one. 

For the final candidate point sets we increase the accuracy of our star discrepancy estimation by performing an additional 50 runs of the TA-algorithm.
These values either re-affirmed the previously computed ones, or they give a slightly larger bound on the star discrepancy of the point set, deviating by at most 5 to 10$\%$ from the previous values.

We note that for all reference point sets for which we know the exact discrepancy, we also do an exact DEM-evaluation of our candidate point set. Only for combinations of $n$ and $d$ where no exact discrepancy values are found in the literature, we resort to the lower bounds provided by the TA-algorithm. These lower bounds are at least as good as the ones presented in the literature, since the new TA-evaluation algorithm is better than the ones used in the previous works, cf.~\cite{GWW12}.

\subsection{Selection}

When dealing with the star discrepancy question, we use tournaments as selection mechanism \cite{Goldberg1991}.
Tournament selection selects the best individual among $k$ individuals which are selected at random from the entire population.

For the inverse star discrepancy question, we are dealing with two competing objectives: the number of points $n$ on the one hand, and the smallest star discrepancy of any $n$-point set on the other. We thus need to resort here to a totally different evaluation and selection scheme. 
We have used for this problem the NSGA-II~\cite{Deb2002}, a standard multiobjective selection algorithm.
The evaluation is made using a bisection method. For a given configuration, we try to find the smallest number of points for which the discrepancy is lower than the threshold. This method allows us to run only $\log(b-a)$ times the discrepancy algorithm for a single configuration, where $a$ and $b$ are the frontiers of the search for the number of points. The bisection evaluation returns both the discrepancy (which is lower than the threshold) and the number of points. 
Here again the discrepancy is either computed by the DEM-algorithm (wherever feasible) or by the TA-heuristic (in all other cases).

\section{Results}
\label{sec:results}

We give a brief overview of the experimental setup in Section~\ref{sec:setup}. We then present the results of our algorithm. For the star discrepancy question, we compare the results obtained by our algorithm with those found in the literature (Sec.~\ref{comp:discrepancy}). 
For the inverse discrepancy problem, we are not aware of any paper addressing this question experimentally. We thus present our results for answering a question on the inverse discrepancy presented in~\cite{NW10} and~\cite{Hinrichs13} (Sec.~\ref{comp:inverse}).

All point configurations achieving the bounds presented below are available online on~\cite{url}. 

\subsection{Experimental Setup}
\label{sec:setup}
All the experiments were run with the algorithm described in the last section. Specific parameters for the operators are given in Table~\ref{tab:params}. Given the satisfying results of these settings, we did not attempt to fine tune these parameters. During each discrepancy experiment the 25 best individuals are kept in an archive. 
Where discrepancy values are computed only by the TA-algorithm and are thus just lower bounds for the exact value, the fitness of the individuals kept in the archive is dynamic as well (i.e., if the value of the individual changes due to a reevaluation, it also changes in the archive). The process is the same for the inverse discrepancy experiments, but instead of a list of the best individuals, the archive contains all the individuals that are not Pareto-dominated by the archive.\footnote{We recall that an element $x \in \R^d$ is Pareto-dominated by a set $S \subseteq \R^d$ if there exists a vector $y\in S$ that for all objectives $i \in \{1, \ldots, d\}$ is at least as good as $x$ (i.e., $y_i \leq x_i$ for minimization objectives and $y_i \geq x_i$ for objectives to be maximized), and is strictly better ($y_i < x_i$ and $y_i >x_i$, respectively) for at least one objective $i\in \{1, \ldots, d\}$.}

At the end of an experiment, we send to the final evaluation the entire archive and the final population.

\begin{table}
	\centering
	\begin{tabular}{l|c|c|c} \hline
		Dimensionality		& 4 -- 10			& 11 -- 25			& 100	\\ \hline
		Generations			& 50					& 100				& 200				\\
		Population Size	& $(25 + 100)$	& $(25 + 100)$	& $(25 + 100)$	\\
		Crossover Prob.	& $0.7$				& $0.7$				& $0.7$				\\
		Mutation Prob.		& $0.3$				& $0.3$				& $0.3$				\\
		Match Prob Mut.	& $0.05$			& $0.05$			& $0.05$			\\
		Tournament Size	& 3					& 3					& 3					\\ \hline
	\end{tabular}
	\caption{Optimization algorithm parameters.}
	\label{tab:params}
\end{table}

\subsection{Low Discrepancy Point Sets}
\label{comp:discrepancy} 

Tables~\ref{tab:res_dgw_exact} to~\ref{tab:res_thi_approx} compare the discrepancy values obtained with the genetic optimization of the Halton sequence against the results presented in~\cite{Thi01JoC, Thi01b}, and~\cite{DGW10}. Except for $d=4$ and $n=625$,
all our point sets achieve a lower discrepancy than what has been presented in the literature. In fact, we can observe that our sequences present sometimes a discrepancy that is only half as large as the bounds presented in the literature.

\begin{table}
	\centering
	\begin{tabular}{l|c|c|c} \hline
		$d$		& $n$		& Results from \cite{DGW10}	& Optimized Halton	\\ \hline
		5			& 95			& $\sim0.11$					& $0.08445$		\\ \hline
		7			& 65			& $0.150$					& $0.1361$		\\
		7			& 145		& $0.098$					& $0.08640$		\\ \hline
		9			& 85			& $0.170$					& $0.1435$		\\ \hline
	\end{tabular}
	\caption{Exact discrepancy results for the sequences presented in \cite{DGW10}.}
	\label{tab:res_dgw_exact}
\end{table}

\begin{table}
	\centering
	\begin{tabular}{l|c|c|c} \hline
		$d$		& $n$		& Results from \cite{DGW10}	& Optimized Halton	\\ \hline
		9*			& 145		& $0.119$					& $0.1083$		\\ \hline
		12*		& 65			& $0.276$					& $0.2049$		\\
		12			& 145		& $0.156$					& $0.1354$		\\ \hline
		15			& 65			& $0.322$					& $0.2413$		\\
		15			& 95			& $0.258$					& $0.1969$		\\
		15			& 145		& $0.198$					& $0.1589$		\\ \hline
		18			& 95			& $0.293$					& $0.2237$		\\
		18			& 145		& $0.230$					& $0.1823$		\\ \hline
		20			& 145		& $0.239$					& $0.1947$		\\ \hline
		21			& 95			& $0.299$					& $0.2434$		\\ \hline
	\end{tabular}
	\caption{Approximated discrepancy results for the sequences presented in \cite{DGW10}. Lines marked with a star indicate that our final discrepancy measure is exact.}
	\label{tab:res_dgw_approx}
\end{table}

\begin{table}
	\centering
	\begin{tabular}{l|c|c|c} \hline
		$d$		& $n$		& Results from \cite{Thi01JoC, Thi01b}	& Optimized Halton	\\ \hline
		4			& 125		& $0.089387$			& $0.05609$		\\
		4			& 625		& $0.01772458$		& $0.01905$		\\ \hline
		5			& 25			& $0.238297$			& $0.1800$		\\
		5			& 125		& $0.1417881$			& $0.07158$		\\
		5			& 625		& $0.02666228$			& $0.02352$		\\ \hline
		6			& 49			& $0.210972$			& $0.13959$		\\ 
		6			& 343		& $0.08988426$		& $0.04547$		\\ \hline
		7			& 49			& $0.2690111$			& $0.1641$		\\ \hline
		8			& 121		& $0.1701839$			& $0.1090$		\\ \hline
		9			& 121		& $0.2121262$			& $0.1244$		\\ \hline
	\end{tabular}
	\caption{Exact discrepancy results for the sequences presented in \cite{Thi01JoC,Thi01b}.}
	\label{tab:res_thi_exact}
\end{table}

\begin{table}
	\centering
	\begin{tabular}{l|c|c|c} \hline
		$d$		& $n$		& Results from \cite{Thi01JoC,Thi01b}	& Optimized Halton	\\ \hline
		7*			& 343		& $0.129832$					& $0.05192$		\\
		7			& 2401		& $0.030518$					& $0.01518$		\\ \hline
		10*		& 121		& $0.2574323$					& $0.1334$		\\
		10			& 1331		& $0.093028$					& $0.03251$		\\ \hline
		11*		& 121		& $0.301048$					& $0.1402$		\\ \hline
		12			& 169		& $0.271837$					& $0.1211$		\\
		12			& 2197		& $0.096713$					& $0.02857$		\\ \hline
		15			& 289		& $0.256021$					& $0.1083$		\\
		15			& 4913		& $0.085855$					& $0.02239$		\\ \hline
		20			& 529		& $0.259366$					& $0.09859$		\\ \hline
		100		& 101		& $0.954159$					& $0.5458$		\\ \hline
	\end{tabular}
	\caption{Approximated discrepancy results for the sequences presented in \cite{Thi01JoC,Thi01b}. Lines marked with a star indicate that our final discrepancy measure is exact.}
	\label{tab:res_thi_approx}
\end{table}

\begin{figure}
	\centering
	\includegraphics[width=0.95\linewidth]{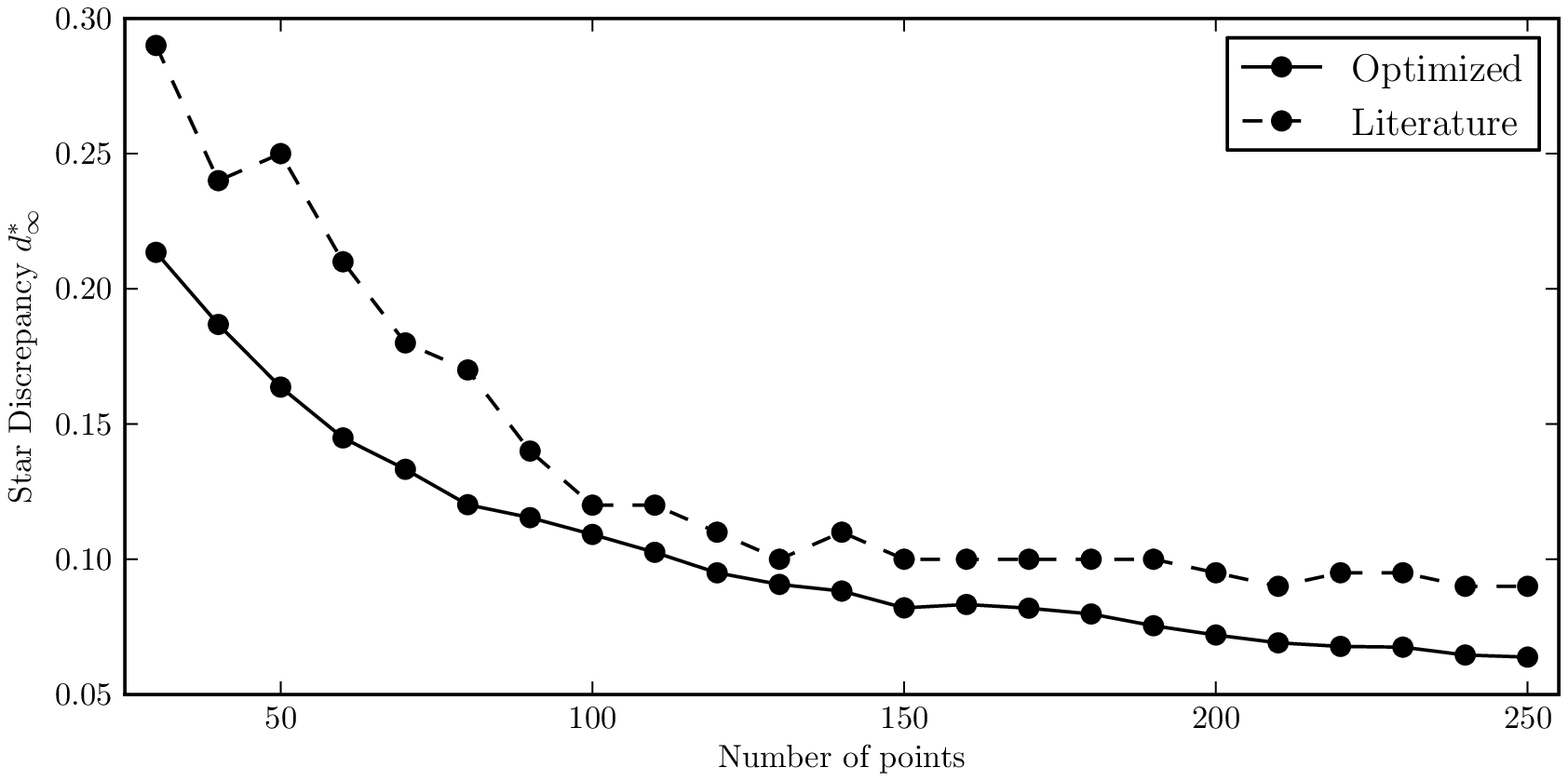}
	\caption{Exact discrepancy results on 7 dimension point sets.}
	\label{fig:7d}
\end{figure}

Figure~\ref{fig:7d} presents the discrepancy of our optimized point sets against an aggregation of the best point sets in 7 dimensions of \cite[Figure 5]{Thi01b}. Again we see that all our values are smaller than those presented there.

\begin{figure}
\begin{center}
\includegraphics[width=0.95\linewidth]{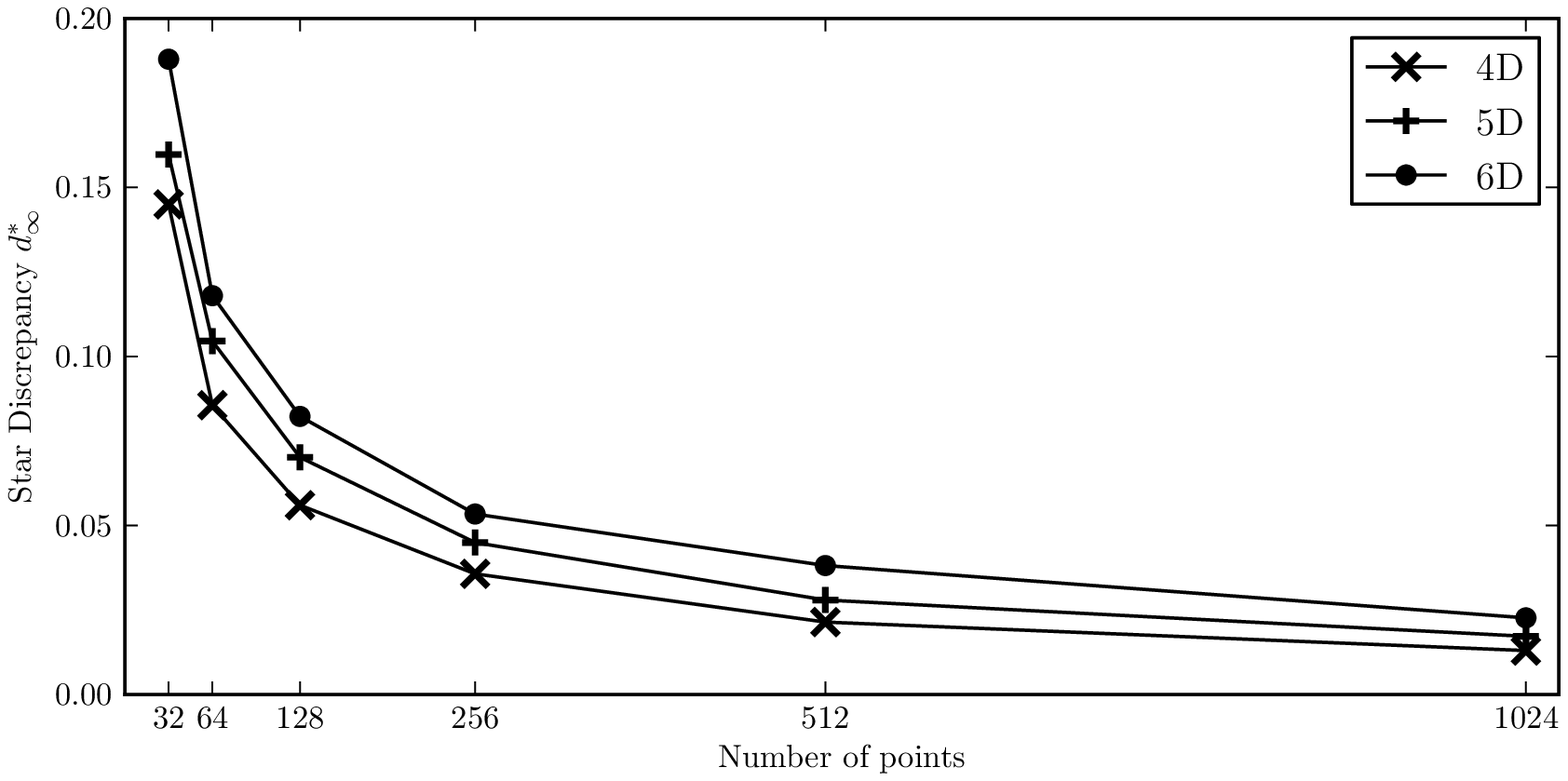}
\end{center}
\caption{Best exact star discrepancy values in $d=4$ to $d=6$.}
\label{GRAPH2}
\end{figure}

\begin{figure}
\begin{center}
\includegraphics[width=0.95\linewidth]{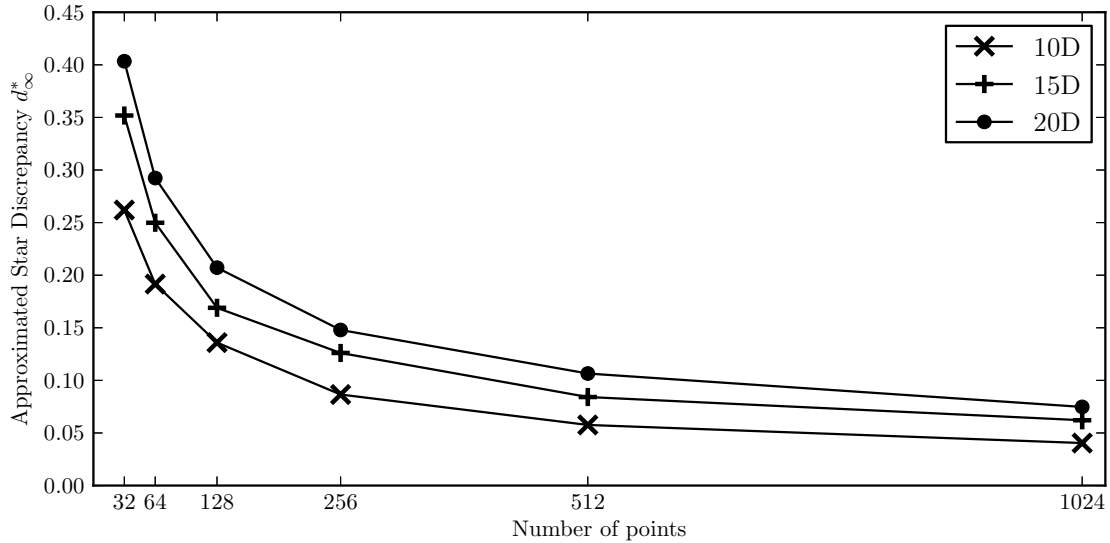}
\end{center}
\caption{Best approximated star discrepancy values in $d=10$ to $d=20$.}
\label{GRAPH3}
\end{figure}

The efficiency of our algorithm allows us to do much more than what can be represented by the comparisons with previous work. As an example, we present in the following our results for minimizing the discrepancy in fixed dimension for $n$ ranging from 32 to 1024 points, cf.~Figures~\ref{GRAPH2} and~\ref{GRAPH3}.

Figure~\ref{GRAPH2} shows for $4, 5$ and $6$ dimensions how the star discrepancy decreases with growing $n$. The values computed in this figure are exact values, i.e., they are computed with the DEM-algorithm. As expected, we can see that with growing dimension, more points are needed to achieve the same discrepancy bounds.

In Figure~\ref{GRAPH3} we plot similar results for dimension in which the DEM-algorithm is infeasible. The values plotted in this chart are thus computed by 
50 runs of 
the TA-algorithm. 
The situation in such higher dimensions seems to be very similar to that in the smaller ones.

These plots could be easily extended to more points and to higher dimensions, showing that we can easily get, for any combination of $d$ and $n$, a point configuration of low star discrepancy value.

\subsection{Inverse Star Discrepancy}
\label{comp:inverse}

The inverse star discrepancy problem is even more complex than the star discrepancy one, since it has an additional parameter, which is the number of points to be placed in the $d$-dimensional unit cube. Given the computational intractability of star discrepancy evaluation, it is therefore natural that not much empirical work has been done to address this question. 

Our algorithm, however, can be easily adjusted to compute inverse star discrepancies, as explained in~Sec.~\ref{sec:algorithm}. We use this approach to address Open Problem 42 in~\cite{NW10}. 
The first subproblem asks to construct a point set $X$ in 15 dimensions such that $\disc(X) \le 0.25$ and $|X|\leq 1528$. 
By an approach suggested by Hinrichs~\cite{Hinrichs13}, it suffices to construct an $8$-dimensional point set $X'$ of star discrepancy at most $0.125$ and size $|X'|\leq 764$. 
Using his \emph{lifting technique}, $X'$ can be turned into a 15-dimensional set $X$ of size $2|X'|$ and discrepancy at most $2\disc(X')$.
We are thus interested in the inverse star discrepancy problem with $d=8$ and $\varepsilon=0.125$. 
By similar arguments (see~\cite{Hinrichs13} for the details), we can solve the 15-dimensional problem also by finding a 
4-dimensional point configuration of star discrepancy at most $0.0136$ or a 5-dimensional point configuration of discrepancy at most $0.0575$ with at most $764$ points each. 
As we can see from Table~\ref{tab:inv_discr}, our algorithm easily solves this 15-dimensional problem. For $d=8$, it outputs a point configuration of discrepancy $0.1248$ and size $n=104$. This is much less than the requested $764$ size bound asked for in~\cite{NW10,Hinrichs13}. Also the 5-dimensional version of the problem is solved easily; the point configuration has only 172 instead of the requested upper bound of $764$ points. Only for the 4-dimensional one the lifting approach of Hinrichs seems too weak to solve the 15-dimensional problem. 

We see a similar behavior for the other two subproblems in~\cite{NW10}: the original problems (find, (i), a point configuration $X$ in $d=30$ of discrepancy $\disc(X)\leq 0.25$ and size $|X| \leq 3187$ and, (ii), a configuration $X'$ in $d=50$ dimensions with $\disc(X')\leq 0.25$ and $|X'|\leq 5517$) can be solved easily for the first two steps of the lifting technique (15 and 10 dimensions, and 25 and 17 dimensions, respectively), but for the last reported step (in 8 and 13 dimensions, respectively) we do not find point configurations meeting the required bound. It is therefore very likely that generalized Halton point sets with such small discrepancy values do not exist. 

Table~\ref{tab:inv_discr} is to be read as follows. The original three subproblems of~\cite[Open Problem 42]{NW10} are the ones in bold red print.
The expected (Exp.~$n$) column presents the number of points required by the problem and by Hinrichs' method, respectively. 
The bounds column shows the selected search space for the bisection algorithm. The upper bounds were selected from preliminary experiments with randomly scrambled Halton sequences. The lower bounds have been set so that the maximum number of trials in the bisection search is not too high. 
As mentioned above, it can be seen that for all problems our algorithm finds a suitable sequence that answers the open problems. 
We did not complete the computations for $d=8$ and $d=13$ and $\varepsilon=0.0136$ since the bounds for these instances are already much larger than what is requested. 
The starred values in the table are computed by 50 individual runs of the TA-algorithm. Before officially declaring~\cite[Open Problem 42]{NW10} solved, the starred values need to be re-affirmed by an exact algorithm (note that the runtime of the DEM-algorithm would be several years on these instances).

\begin{table}
	\centering
	\begin{tabular}{l|l|c|c|c|c} \hline
		$d$		& $\varepsilon$	& Exp.~$n$	& Bounds	& $n$	& $\disc(X')$	\\ \hline
			\textbf{\textcolor{red}{15}}			& \textbf{\textcolor{red}{$0.25$}}			& \textbf{\textcolor{red}{$\leq 1528$}}	&&& 	\\
				8			& $0.125$				& $\leq 746$			& 	[64,128]	& 104	& $0.1248$	\\
		5			& $0.0575$			& $\leq 746$			& 	[128,256]	& 172	& $0.0573$	\\
			4			& $0.0136$			& $\leq 746$			& 	[1044,1300]	& 1048	& $0.0135$	\\
 \hline
 	\textbf{\textcolor{red}{30}}			& \textbf{\textcolor{red}{$0.25$}}			& \textbf{\textcolor{red}{$\leq 3187$}}	&&& 	\\
 			15*		& $0.125$				& $\leq 1593$		& 	[152,280]	& 251	&	$0.1245$	\\
 					10*		& $0.0575$			& $\leq 1593$		& 	[472,600]	& 537	& $0.0574$	\\
		8*			& $0.0136$			& $\leq 1593$		& 	-				& $>3000$	& -	\\
 \hline
 \textbf{\textcolor{red}{50}}			& \textbf{\textcolor{red}{$0.25$}}			& \textbf{\textcolor{red}{$\leq 5517$}}	&&& 	\\
 	25*		& $0.125$				& $\leq 2758$		& 	[422,550]	& 513	&	$0.1245$	\\
 	17*		& $0.0575$			& $\leq 2758$		& 	[1094,1350]	& 1239	& $0.0575$	\\
	13*		& $0.0136$			& $\leq 2758$		& 	-				& $>5000$	& -	\\

 \hline
	\end{tabular}
	\caption{Inverse discrepancy results for the three problem instances. Final discrepancy of stared lines are approximated.}
	\label{tab:inv_discr}
\end{table}

\begin{figure}
	\centering
	\includegraphics[width=0.95\linewidth]{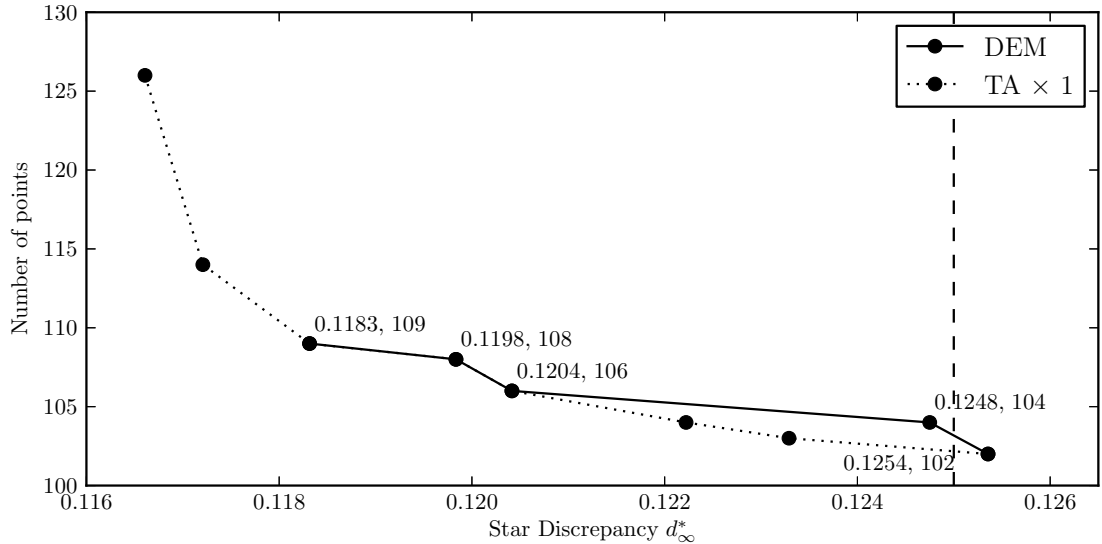}
	\caption{8 dimensional trade-offs between the number of points and the discrepancy found.}
	\label{fig:tradeoff8d}
\end{figure}

\begin{figure}
	\centering
	\includegraphics[width=0.95\linewidth]{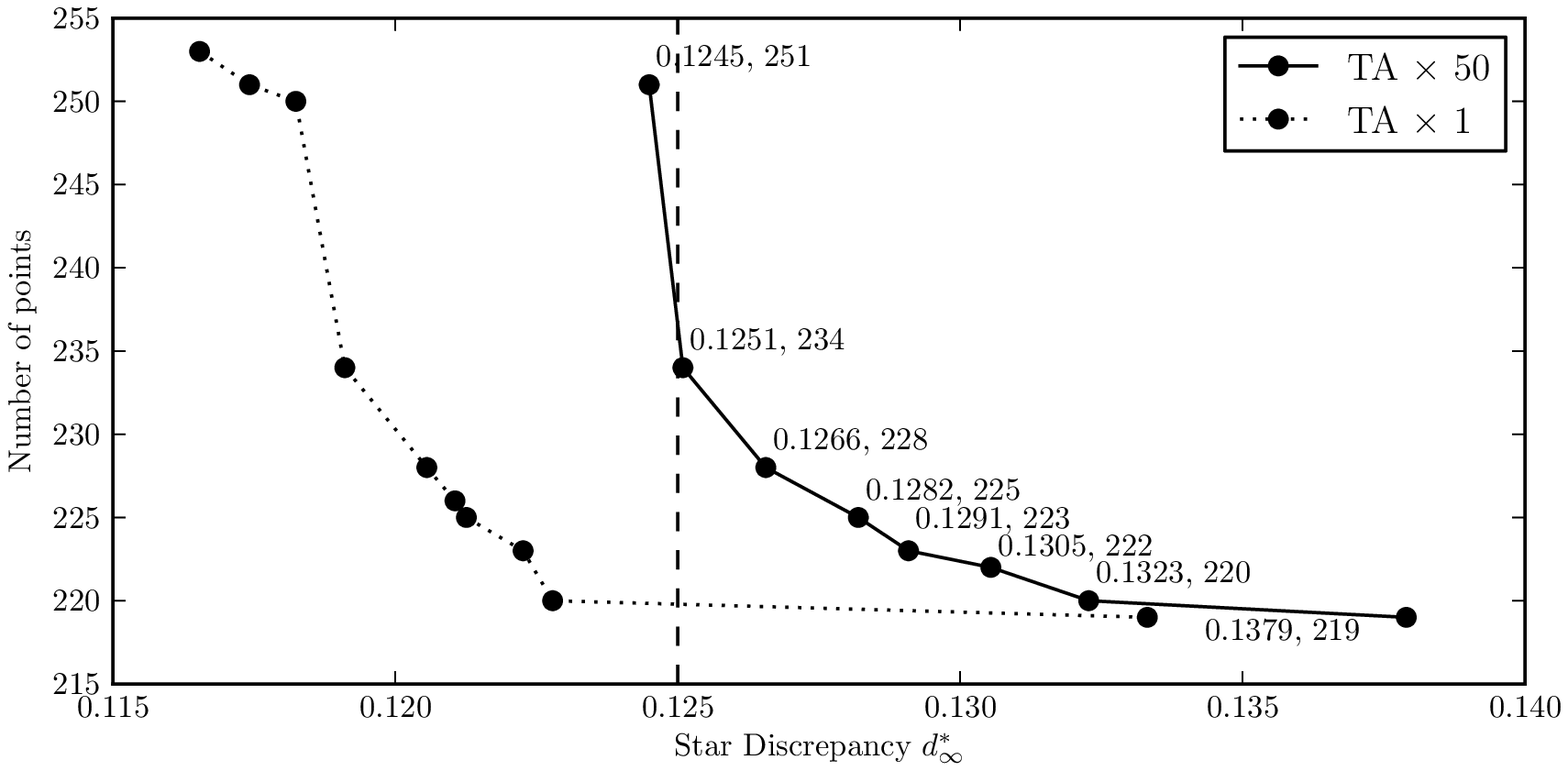}
	\caption{15 dimensional trade-off between the number of points and the discrepancy.}
	\label{fig:tradeoff15d}
\end{figure}

\begin{figure}
	\centering
	\includegraphics[width=0.95\linewidth]{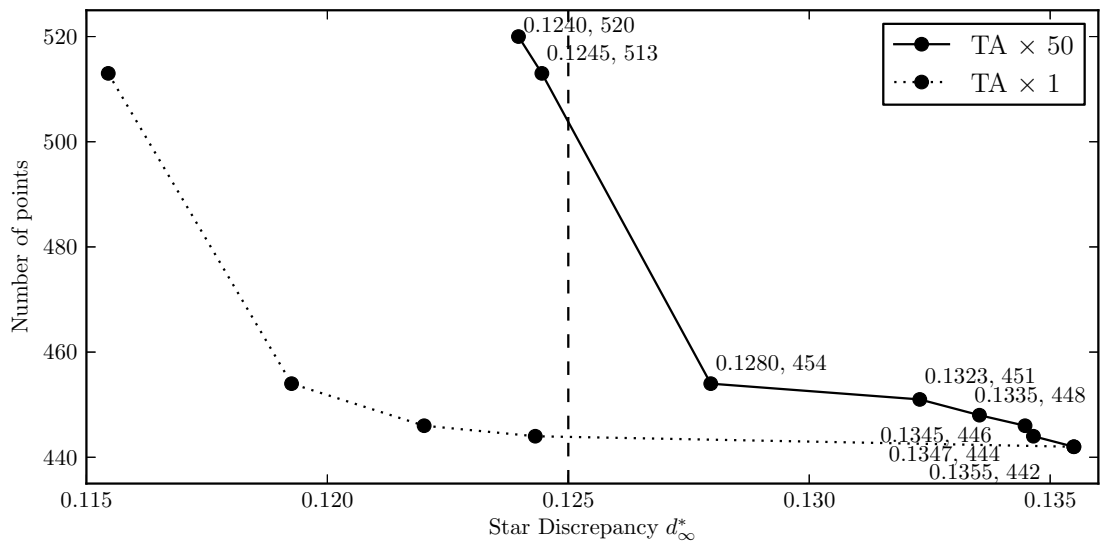}
	\caption{25 dimensional trade-off between the number of points and the discrepancy.}
	\label{fig:tradeoff25d}
\end{figure}

Figures~\ref{fig:tradeoff8d} to \ref{fig:tradeoff25d} present the Pareto front of the final populations for the 8-, 15-, and 25-dimensional cases.\footnote{Note here that the axis are switched, as the objective here is to minimize, for a given maximal star discrepancy value, the number of points $n$.} It can be seen that our method achieves a nice diversity in the found point sets giving a broad range of candidates with different trade-offs. The difference between the final front and the reevaluated one can be explained by the fact that the 
optimization algorithm takes advantage of the errors made by the TA-algorithm. For example, the same sequence can appear multiple times during the evolution, most of the time its approximated discrepancy will be very close to the true one, but it takes only one bad evaluation (the approximation is lower than expected) for that point set to make its way into the archive. Increasing the number of repetitions of the TA-algorithm would prevent this from happening, but it would also increase the running time of the optimization.

\section{Conclusions}

We have presented a new algorithm for computing low star discrepancy point sets. As shown in Section~\ref{sec:results}, our results outperform previous point configurations. Furthermore, our algorithm can be easily adapted to compute upper bounds for the inverse discrepancy. 
The point sets are available online at~\cite{url}. We are confident that our algorithm and the generated point sets will be useful in a broad range of applications. Most notably, our point sets can ensure a better approximation errors in quasi-Monte Carlo numerical integration and in experimental design.

It would be interesting to study whether our results can be further improved by resorting to other point sets, such as (scrambled) Sobol or Faure configurations.

Another interesting research direction concerns the evaluation of 
star discrepancy values. As exhibited above, this is a provably difficult task. Still we have seen that the TA-algorithm from~\cite{GWW12} provides an accurate estimate wherever this can be checked. Is it possible to design similar heuristics for computing reasonable \emph{upper bounds} for the star discrepancy of a given point set? 

\newpage
\section*{Acknowledgments}
 We would like to thank Magnus Wahlstr\"om from the Max Planck Institute for Informatics for providing an implementation of the DEM algorithm~\cite{DEM96}, for his excellent support regarding the evaluation algorithm~\cite{GWW12}, and for his constructive feedback regarding our work.

We would also like to thank Christian Gagn\'e and Michael Gnewuch for several very helpful discussions on the topic of this work.

Carola Doerr is supported by a Feodor Lynen postdoctoral research fellowship of the Alexander von Humboldt Foundation and by the Agence Nationale de la Recherche under the project ANR-09-JCJC-0067-01.

Fran\c cois-Michel De Rainville is supported by the FQRNT.

All computations have been made possible by the access to Calcul/Compute Qu\'ebec/Canada supercomputing facilities.

%\bibliographystyle{amsalpha}
%\bibliography{biblio}
\providecommand{\bysame}{\leavevmode\hbox to3em{\hrulefill}\thinspace}
\providecommand{\MR}{\relax\ifhmode\unskip\space\fi MR }
\providecommand{\MRhref}[2]{%
  \href{http://www.ams.org/mathscinet-getitem?mr=#1}{#2}
}
\providecommand{\href}[2]{#2}

\end{document}